\pdfoutput=1

\documentclass[11pt]{article}

\usepackage[final]{acl}

\usepackage{times}
\usepackage{latexsym}

\usepackage[T1]{fontenc}

\usepackage[utf8]{inputenc}

\usepackage{times}
\usepackage{latexsym}
\usepackage{booktabs}
\usepackage{amsmath}
\usepackage{float}
\usepackage{listings}
\usepackage{adjustbox}
\usepackage{longtable}
\usepackage{xcolor,colortbl}
\usepackage{diagbox} 
\usepackage{array,multirow}
\usepackage{graphicx} 
\usepackage{placeins} 
\usepackage{microtype} 
\usepackage{enumitem} 
\usepackage{amsmath, bm} 
\usepackage{comment} 
\usepackage{amsmath, bm, amssymb}
\usepackage{subcaption}
\usepackage{transparent}
\usepackage{geometry}
\usepackage{pdflscape}
\usepackage{tcolorbox}

\usepackage{booktabs}
\usepackage{tabularx}
\usepackage{xltabular}
\usepackage{fvextra}

\tcbuselibrary{listings, breakable}
\definecolor{grey}{named}{gray}
\usepackage{listings}
\usepackage{hyperref}
\usepackage{url}
\usepackage{xurl}

\usepackage{microtype}

\usepackage{inconsolata}

\definecolor{mF}{HTML}{A9D18E} 
\definecolor{mNDCG}{HTML}{5187ED} 
\definecolor{mQ}{HTML}{FFD966} 


\title{Robustness of Chatbot LLM Judgments of Presidential Actions and Hypothetical Reforms}

\author{Oleg Vasilyev \\
  Independent Researcher \\
  \texttt{{vasilyev.oleg@gmail.com}}\\}

\begin{document}
\maketitle
\begin{abstract}
Chatbots are increasingly consulted about public affairs, and their judgments of policies are useful only if they are robust. We measure this robustness in the setting of an ordinary user, through consumer chat interfaces. For 30 days, five chatbot configurations (Mistral, Gemini Flash, Claude Sonnet, and Claude Opus at two effort settings) were asked twice a day to score the long-term effect on the United States of 40 actions of the last four presidential administrations and of 15 hypothetical reforms, each with a confidence and a short comment. The judgments are robust: all configurations agree on the sign and the ranking of the administrations' mean scores, and two different configurations asked on the same day give opposite signs to the same action in about 4\% of cases. Responses from different days differ more than same-day responses (which shared the order of the items); the difference grows slowly over the month, except where the service itself changed. We identify three such changes, which shifted the judgments while leaving little or no trace in the interface. Treating the comments as a retrieval task, we find them specific to each item and consistent across runs and chatbots, with lexical and embedding similarities agreeing. We release the dataset of 16{,}500 judgments.
\end{abstract}

\section{Introduction}\label{sec:Introduction}
Chatbots are becoming a common source of information on public affairs \citep{reuters2026dnr, bruinsma2026issuebiasgenerativeai}. AI systems have also been shown to help people find common ground in political deliberation \citep{tessler2024ai, peters2026polistemicsevaluatingllmsinformation}. Judging high-level human decisions with AI may be helpful for two reasons:
\begin{enumerate}[topsep=0pt,itemsep=-1ex,partopsep=1ex,parsep=1ex]
    \item The judgment is mediated by a model and by the accumulated human knowledge it has learned, rather than driven by immediate political stimuli and the emotional impulses of a specific person or a small group of people.
    \item The judgment is fast and can cover many factors and possible consequences.
\end{enumerate}

Such judgments are useful only if they are robust. It is therefore important to estimate, at least empirically and crudely, how robust AI judgments of complex human decisions are. We consider not an evaluation system built around fixed versions of specific models, but the practical setting of a person using a chatbot, who controls neither the model version nor the routing, system, tools or retrieval behind the interface. The practical questions are:
\begin{enumerate}[topsep=0pt,itemsep=-1ex,partopsep=1ex,parsep=1ex]
    \item How robust is a judgment to the choice of chatbot?
    \item How robust is it to drift from day to day, as knowledge and the systems change?
    \item How robust is it within a day, given the randomness that sampling, routing and tool use bring in?
    \item Does robustness differ between judging decisions already implemented and judging hypothetical ones?
\end{enumerate}

\paragraph{Political views of LLMs.}
Most studies of LLMs and politics measure a model's own positions with opinion surveys and political questionnaires, and typically find left-of-center leanings \citep{pmlr-v202-santurkar23a,rozado2024political}. Such studies compare a model's opinions with those of a population or of its subgroups, but which group, if any, a model should agree with is unclear, as \citet{pmlr-v202-santurkar23a} themselves note. Moreover, the robustness of such measurements has been questioned: positions elicited with forced-choice questions change when models may respond freely \citep{rottger-etal-2024-political}, and the abstract positions of models do not predict how they vote on concrete referenda \citep{barmettler2026instrument}. On the other hand, responses are fairly consistent across paraphrases, formats and languages, though less so on controversial topics \citep{moore-etal-2024-large}, and stable across repeated runs of the same questionnaire \citep{sakhawat2026political,hanna2026stability}. Here we ask not for agreement with statements but for quantified judgments of the consequences of specific government actions, real and hypothetical.

\paragraph{Change over time.}
\citet{chen2024chatgpt} showed that the behavior of GPT-3.5 and GPT-4 changed substantially between their March and June 2023 versions. \citet{cen2025longitudinal} queried twelve models almost daily during the 2024 US election season and found both gradual drift and abrupt steps, some coinciding with known model updates and others unexplained. Over 36 weeks of weekly queries, \citet{nogueira2026changing} found a small drift to the right in the political responses of ChatGPT, Gemini and Claude, often coinciding with new releases, together with more hedging over time; \citet{aksoy2026evaluating} propose a framework for tracking such biases across model releases.

\paragraph{Run-to-run variation and chat interfaces.}
Even at temperature zero, LLMs served through APIs give different outputs when a query is repeated, with accuracy varying by up to 15\% between runs \citep{atil-etal-2025-non}. A chat interface adds further unknowns, since the model version, system prompt, routing and retrieval are chosen by the provider and may change without notice. In an audit of three chatbots answering prompts based on disinformation about the war in Ukraine through their standard interfaces, the same prompt submitted in new chats gave responses of different accuracy for up to 53\% of the prompts \citep{makhortykh2024stochastic}.

\paragraph{LLMs as judges.}
LLMs are widely used to judge the quality of texts, where their verdicts agree well with human preferences but show position, verbosity and self-enhancement biases \citep{zheng2023judging}. In our case the object of judgment is a real or hypothetical government decision, and no reference response exists, so robustness is the property that can be measured.

\begin{figure}[t]
    \centering
    \includegraphics[width=\columnwidth]{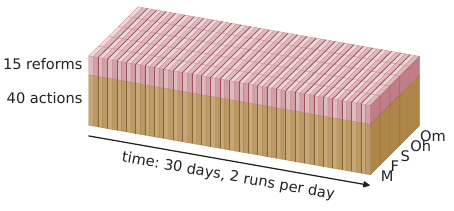}
    \caption{All 600 prompt runs: 5 chatbot configurations $\times$ 2 prompts $\times$ 30 days $\times$ 2 runs a day, shown as small boxes. The setup is explained in Section~\ref{sec:Setup}.}
    \label{fig:runs_block}
\end{figure}

\paragraph{This work.}
To answer the questions above in the setting of an ordinary user, we queried five chatbot configurations from three providers through their consumer chat interfaces every day for 30 days, from September 13 to October 12, 2026, in the weeks before the US midterm elections. Each day, each chatbot judged 40 real US presidential actions, from the Obama administration to the second Trump administration, and 15 hypothetical US government reforms, giving for each item a score, a confidence and a short comment. Each prompt was run twice a day, each time in a fresh temporary chat, the second time immediately after the first, which gives the same-day difference. Figure~\ref{fig:runs_block} illustrates all 600 prompt runs; together they contain 16{,}500 judgments.

Our contributions are:
\begin{enumerate}[topsep=0pt,itemsep=-1ex,partopsep=1ex,parsep=1ex]
    \item A public dataset of 16{,}500 judgments of 40 presidential actions and 15 hypothetical reforms by five chatbot configurations, collected twice a day over 30 days, each with a score, a confidence and a comment explaining the score\footnote{\url{https://huggingface.co/datasets/oleg-vasilyev/llm-policy-judgments}}.
    \item Evidence that the judgments are robust to the choice of chatbot and the day: all configurations agree on the sign of each administration's mean score and on the ranking of the administrations; two different configurations asked on the same day give opposite signs to the same action in only about 4\% of cases.
    \item A comparison of the same-day noise, the day-to-day drift and the differences between chatbots; the identification of three changes in the services behind the chat interfaces during the month (the changes shifted the judgments while leaving little or no trace in the interface).
    \item A use of standard retrieval measures to evaluate the consistency of the comments that explain the judgments. The comments are specific to each item and consistent across runs and chatbots; the conclusions agree across lexical and embedding similarity and across ranking measures. For the hypothetical reforms, word overlap identifies the comments across chatbots better than embeddings do, suggesting that specific terminology carries much of the distinction.
\end{enumerate}

\section{Setup}\label{sec:Setup}
\subsection{Items to Judge}\label{ssec:Items}
The presidential actions were selected with the help of Claude Opus 5, which we asked for the 40 most consequential presidential actions of the four latest US administrations, from the Obama administration to the beginning of 2026. Ending the selection there kept most of the actions within the training data of the models, and left more than half a year between the latest action (January 2026) and the start of the experiment.

\begin{figure}[t]
    \centering
    \includegraphics[width=\columnwidth]{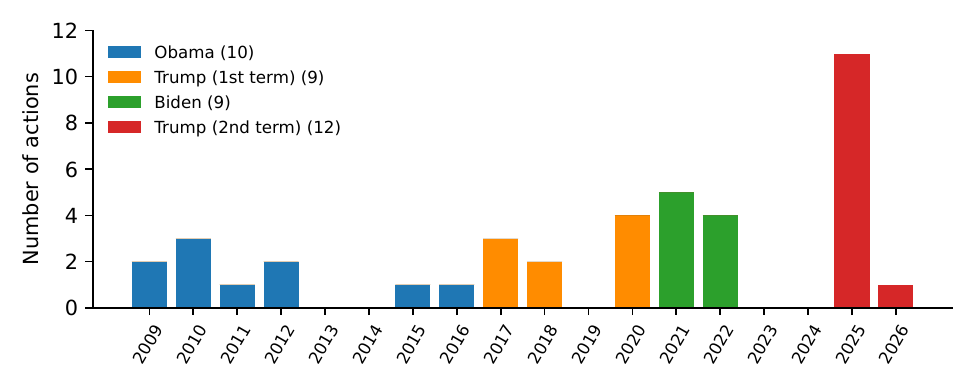}
    \caption{The 40 presidential actions by the year in which each was launched (the first year, for actions that span several years). By administration: Obama 10, Trump1 9, Biden 9, Trump2 12.}
    \label{fig:ActionsPerYear}
\end{figure}

The actions are listed with their descriptions in Table~\ref{tab:p2-actions} (Appendix~\ref{sapp:items_actions}), and their distribution over years and administrations is shown in Figure~\ref{fig:ActionsPerYear}.

For more challenging judgments, we also defined, with the selection guided by Claude Opus 5, 15 hypothetical US government reforms intended to be among the most beneficial that could be proposed (Table~\ref{tab:f-actions}, Appendix~\ref{sapp:items_reforms}). Six of them reform political institutions and state capacity (I01--I06), eight are economic and social policies, from housing and permitting to education and the fiscal path (I07--I14), and one insures against catastrophic risks, from pandemics to AI (I15).

Some of the reforms are arguably too hypothetical: they are designed to remedy the very conditions that prevent such reforms from being launched. If the government could be motivated to launch them, there might be no need for them. The exercise is therefore a thought experiment: we assume that, for a short interval of time, the motivation of the politicians switches to the benefit of the country, and that the reforms are launched within that interval. The prompt expresses this by asking the models to assume that each reform is taken now and fully supported by all branches of government.

\subsection{Judgment}\label{ssec:Judgment}
The judgment of the actions is defined by the prompt in Figure~\ref{fig:p2-prompt}. It asks for the net effect of each action on the long-term flourishing of the US population over the 50 years after the action was taken. The score is an integer from $-5$ to $+5$: its sign gives the direction of the effect, and its size is anchored to rough numbers of lives, shares of national income, and changes to core institutions. The confidence, an integer from 1 to 5, states how sure the model is of the score, including its sign, and a comment of at most 100 characters gives the reason. The models are asked to judge each action on its own merits, independently of the other actions and of any overall view of the administration, to include consequences that became apparent later, and to base the score on empirical evidence rather than on a political judgment.

The judgment of the reforms is defined by the prompt in Figure~\ref{fig:f-prompt}. It asks for the effect of each reform on human flourishing in the United States over the next 50 years, compared with a baseline of current law and current trends, in expectation rather than as the most likely outcome, and ignoring implementation cost and political difficulty. The score (called \textit{magnitude} in the prompt and in the dataset) is an integer from 1 to 5 and has no sign, because the reforms were designed to be beneficial; a reform judged harmful falls in the lowest band. The bands are anchored to the permanent change in national income, with averted deaths converted at roughly \$10M each, and to the scale of institutional change. Besides the confidence and the comment, the prompt asks for:
\begin{enumerate}[topsep=0pt,itemsep=-1ex,partopsep=1ex,parsep=1ex]
    \item Dimension: the dimension that determined the score: \textit{lives}, \textit{income} or \textit{institutional}.
    \item Irreversibility (1 to 5): how hard the reform would be to undo within 50 years if it proved wrong; 5 means constitutional or effectively permanent, 1 repealable by ordinary legislation.
    \item Contestedness (1 to 5): how much informed experts disagree about the score, independently of the model's own confidence; 5 means deeply contested, 1 not contested.
\end{enumerate}

Both prompts ask only for the effect on the United States, ignoring the rest of the world, because this is common in US policy debate, and it makes the scoring easier. Both prompts also require outpus as one line per item in a fixed format, which was parsed into the fields of the dataset.

\subsection{Chatbots}\label{ssec:models}
Each day from September 13 to October 12, 2026, both prompts were run twice on each of the following chatbot configurations (the short notations are used in the figures):
\begin{enumerate}[topsep=0pt,itemsep=-1ex,partopsep=1ex,parsep=1ex]
    \item M: Mistral's Vibe, Think mode (see Section~\ref{sssec:model_changes_mistral});
    \item F: Gemini Flash 3.6;
    \item S: Claude Sonnet 5, effort High;
    \item Oh: Claude Opus 5, effort High;
    \item Om: Claude Opus 5, effort Max.
\end{enumerate}
Each run was made in a fresh temporary chat, so that no earlier conversation could influence the responses. The items were listed in a random order that changed from day to day; both runs of a day used the same order (the same random seed), so the two runs received identical prompts. The second run of the day followed immediately after the first. When a chatbot refused to score or broke the output format, the run was repeated in a new chat.

Throughout the paper we call these chatbot configurations \textit{models} for simplicity, and number the days from 1 (September 13) to 30 (October 12).

\subsection{Dataset}\label{ssec:Dataset}
The resulting dataset consists of four parts:
\begin{enumerate}[topsep=0pt,itemsep=-1ex,partopsep=1ex,parsep=1ex]
    \item Presidential actions (40): id, administration, year, name, description.
    \item Hypothetical reforms (15): id, name, description.
    \item Scored actions (12{,}000 $= 30$ days $\times$ 5 models $\times$ 2 runs $\times$ 40 actions): day, date, model, run, item id, score, confidence, comment.
    \item Scored reforms (4{,}500): day, date, model, run, item id, score, dimension, irreversibility, contestedness, confidence, comment.
\end{enumerate}
It is described in Appendix~\ref{app:dataset} and published on Hugging Face.\footnote{\url{https://huggingface.co/datasets/oleg-vasilyev/llm-policy-judgments}}

\section{Observations}\label{sec:Observations}
\subsection{Scores by Administration}\label{ssec:admins_avg}
\begin{figure}[t!]
    \centering
    \includegraphics[width=\columnwidth]{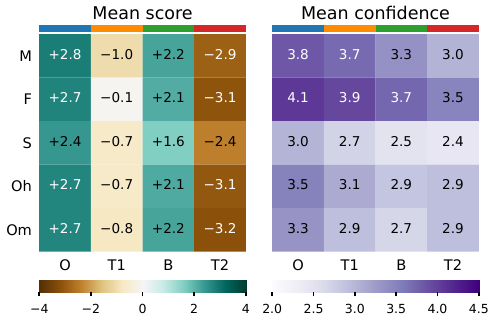}
    \caption{Mean score (left) and mean confidence (right) of each model for the actions of each administration, averaged over all days and runs. O: Obama; T1 and T2: the first and second Trump administrations; B: Biden.}
    \label{fig:ActionsScoresAvg}
\end{figure}

\begin{figure}[t!]
    \centering
    \includegraphics[width=\columnwidth]{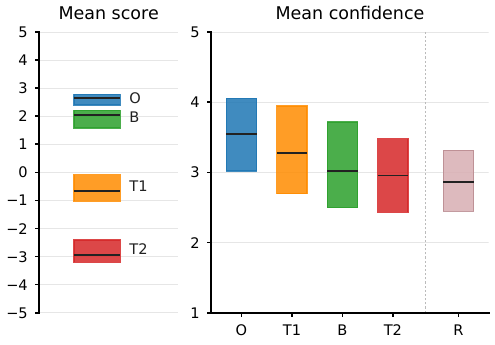}
    \caption{Summary of Figure~\ref{fig:ActionsScoresAvg}. Left: for each administration, the bar spans the lowest and the highest of the five models' mean scores, and the dark line is their average. Right: the same for the confidence, with the confidence in the hypothetical reforms (R) added on the right.}
    \label{fig:p2_admin_ranges}
\end{figure}

The simplest aggregation is by administration and model. Figure~\ref{fig:ActionsScoresAvg} shows each model's mean score and mean confidence for the actions of each administration, averaged over all days and runs. The aggregates are consistent across the models:
\begin{enumerate}[topsep=0pt,itemsep=-1ex,partopsep=1ex,parsep=1ex]
    \item The sign of the mean score is the same for all models: positive for the Obama and Biden administrations, and negative for the two Trump administrations.
    \item All models rank the administrations in the same order: Obama, Biden, the first Trump administration, the second.
    \item The mean confidence is moderate to high for the Obama actions (3.0 to 4.1), the earliest ones, about which the most evidence has accumulated, and falls toward moderate for the later administrations (2.4 to 3.5 for the second Trump administration). The confidence in the judgments of the hypothetical reforms (2.4 to 3.3) is only a little lower than for the actions of the second Trump administration.
\end{enumerate}
Figure~\ref{fig:p2_admin_ranges} shows the same data more concisely. The ranges of the five models' mean scores do not overlap between the administrations; averaged over the models, the mean score is $+2.65$ for Obama, $+2.04$ for Biden, $-0.67$ for the first Trump administration and $-2.95$ for the second. We do not discuss here why the models judge the actions of the Trump administrations, especially the second, so negatively: on the scale of the prompt, a score of $-3$ means a large and durable negative effect. The decline may reflect what \citet{kemp2025goliath} calls the Goliath's Curse. For our purposes it is enough that very different models agree in this scoring.

\subsection{Scores of Individual Items}\label{ssec:noise}
\begin{figure*}[t!]
    \centering
    \includegraphics[width=\textwidth]{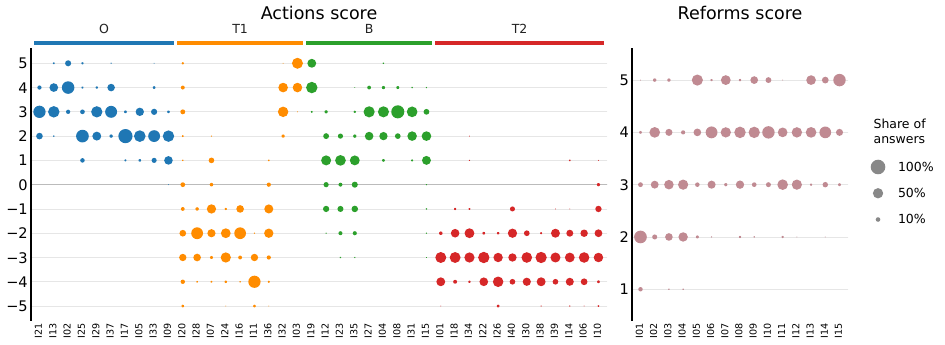}
    \caption{Distribution of the scores of each item over all models, days and runs (300 responses per item). The area of a dot is proportional to the share of the responses. The actions are in chronological order and colored by administration; the reforms are in the order of their ids. The score scales are defined in Figures~\ref{fig:p2-prompt} and~\ref{fig:f-prompt}.}
    \label{fig:item_dots_all}
\end{figure*}

Aggregating only over the models, days and runs, Figure~\ref{fig:item_dots_all} shows the distribution of the 300 scores on each item (5 models $\times$ 30 days $\times$ 2 runs). The scores cluster tightly, although they come from five different models and from 30 days. On average, 94\% of the scores on an action, and 97\% of the scores on a reform, lie within one point of the item's median. The sign of an action's score agrees with the sign of its median in 95\% of the responses. The most divided actions are the reshaping of the federal judiciary (I20), whose scores span the whole scale from $-5$ to $+5$, and the expanded parole and asylum admissions (I35), scored positive in 54\% of the responses, negative in 29\% and zero in 17\%. Figure~\ref{fig:item_dots_models} (Appendix~\ref{app:noise}) shows the distributions for each model separately.

\subsection{Difference between Models}\label{ssec:model_model}
Let $v_M(x,t,i)$ be a numeric field of the response of model $M$ on item $x$ (an action or a reform), on day $t$, in run $i=1,2$: the score, the confidence, or, for the reforms, also the irreversibility or the contestedness. We define the difference between two models $M_1 \neq M_2$ as
\begin{equation}\label{eq:diff_MM_nondiag}
  \frac{1}{4|X||T|} \sum_{x,t,i,j}
  \bigl|v_{M_1}(x,t,i) - v_{M_2}(x,t,j) \bigr|
\end{equation}
Here $|X|$ is the number of the items (40 actions or 15 reforms), $|T|=30$ is the number of the days, and the runs $i,j=1,2$; only the responses of the same day are compared.

The model's own noise is the difference between its two runs of the same day:
\begin{equation}\label{eq:diff_MM_diag}
  \frac{1}{2|X||T|} \sum_{x,t,i\neq j}
  \bigl|v_{M}(x,t,i) - v_{M}(x,t,j) \bigr|
\end{equation}

\begin{figure}[t]
    \centering
    \includegraphics[width=\columnwidth]{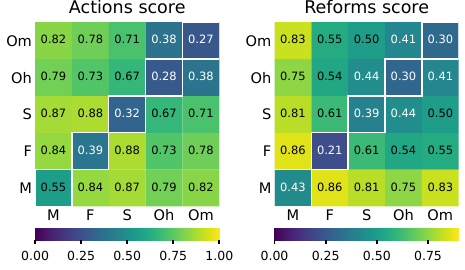}
    \caption{Mean absolute difference between the scores of two models on the same item and day (Equation~\ref{eq:diff_MM_nondiag}, off the diagonal), and between the two runs of the same model (Equation~\ref{eq:diff_MM_diag}, on the diagonal), for the actions (left) and the reforms (right).}
    \label{fig:MM_diff_scores}
\end{figure}

Figure~\ref{fig:MM_diff_scores} shows both for the scores, with Equation~\ref{eq:diff_MM_nondiag} off the diagonal and Equation~\ref{eq:diff_MM_diag} on it. Two runs of the same model on the same day differ by 0.27 to 0.39 points on the action scale for the Claude models and Gemini, and by 0.55 for Mistral. Different models differ by 0.67 to 0.88 points, with one exception: the two Opus settings differ by only 0.38, about as much as each differs from itself. The reform scores show the same structure on their scale from 1 to 5: 0.21 to 0.43 on the diagonal, and 0.41 to 0.86 off it.

\begin{figure*}[t]
    \centering
    \includegraphics[width=\textwidth]{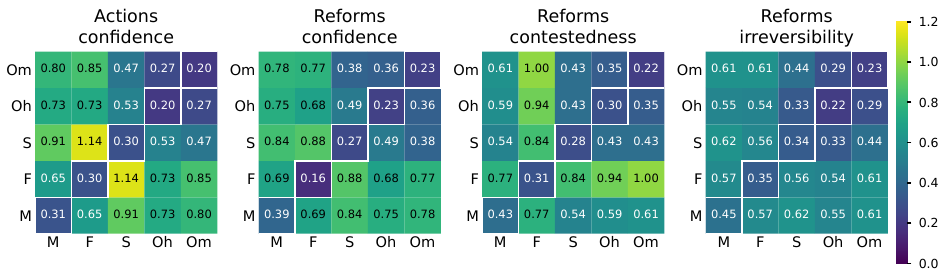}
    \caption{As Figure~\ref{fig:MM_diff_scores}, for the confidence of the actions and for the confidence, contestedness and irreversibility of the reforms.}
    \label{fig:MM_diff_confidence}
\end{figure*}

Figure~\ref{fig:MM_diff_confidence} shows the same for the other fields. For every field, including the scores, the three Claude configurations are closer to each other than to Gemini or Mistral. By administration (Figure~\ref{fig:MM_diff_scores_by_admin}, Appendix~\ref{app:model_model_by_admins}), the models differ least on the Obama actions and most on the actions of the first Trump administration.

\begin{figure}[t]
    \centering
    \includegraphics[width=\columnwidth]{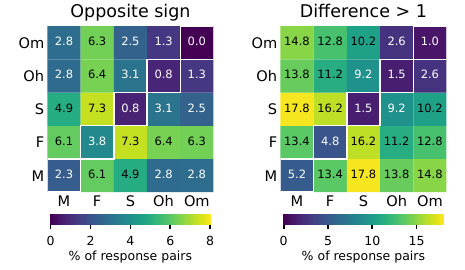}
    \caption{Strong disagreements between responses on the same action on the same day: the share of the pairs of responses with opposite score signs (left), and with scores that differ by more than 1 (right). The pairs are formed as in Equations~\ref{eq:diff_MM_nondiag} and~\ref{eq:diff_MM_diag}; on the diagonal, the two runs of the same model.}
    \label{fig:MM_diff_scores_discrete}
\end{figure}

Figure~\ref{fig:MM_diff_scores_discrete} shows how often two responses on the same action disagree more strongly: in sign, or by more than one point. Encouragingly, two different models asked about the same action on the same day give opposite signs in only 4.3\% of the pairs of responses on average (from 1.3\% to 7.3\% for individual pairs of models), and differ by more than one point in 12\%. Between the two runs of the same model, opposite signs occur in less than 1\% of the pairs for the Claude models, in 2.3\% for Mistral, and in almost 4\% for Gemini, which also disagrees most with the other models.

\subsection{Longitudinal Drift}\label{ssec:drift_scores}
The responses may also drift from day to day. Possible reasons include changes in the routing of a chatbot, updates of the model, and shifts in public knowledge and sentiment, which the chatbots can reach through web search. The period of the experiment, the weeks before the November 2026 midterm elections, may amplify such drift.

For a model $M$, an item $x$ and a number of days $d>0$, we define the drift as
\begin{equation}\label{eq:drift_scores}
  \frac{1}{4|T_d|} \sum_{t,i,j}
  \bigl|v_{M}(x,t+d,i) - v_{M}(x,t,j) \bigr|
\end{equation}
Here $T_d$ is the set of the days $t$ for which $t+d$ is also a day of the experiment ($|T_d|=30-d$), and the other notations are as in Equation~\ref{eq:diff_MM_nondiag}. For $d=0$, only the different runs of the same day are compared:
\begin{equation}\label{eq:drift_scores_0}
  \frac{1}{2|T_0|} \sum_{t,i\neq j}
  \bigl|v_{M}(x,t,i) - v_{M}(x,t,j) \bigr|
\end{equation}
which is Equation~\ref{eq:diff_MM_diag} for a single item. We use $d$ from 0 to 25, because fewer than five pairs of days are left for larger $d$.

\begin{figure*}[t]
    \centering
    \includegraphics[width=\textwidth]{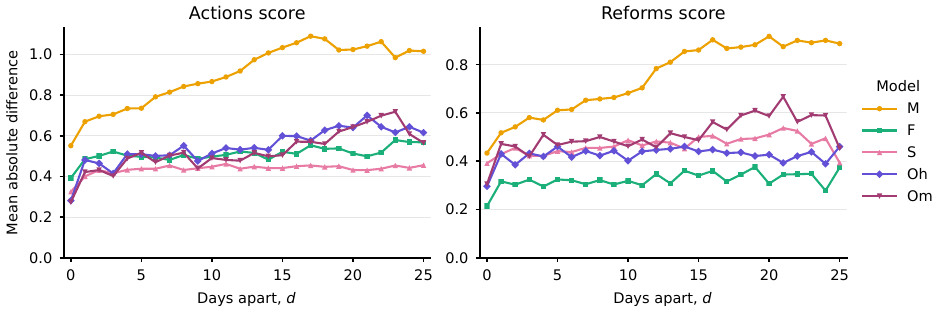}
    \caption{Drift of the scores (Equations~\ref{eq:drift_scores} and~\ref{eq:drift_scores_0}), averaged over the items, as a function of the number of days $d$ between the responses.}
    \label{fig:drift_models}
\end{figure*}

Figure~\ref{fig:drift_models} shows the drift of the scores, averaged over the items. For every model, the responses of consecutive days ($d=1$) differ more than the two runs of the same day ($d=0$); for the actions, by 0.08 to 0.20 points. 
The two runs of a day received identical prompts, while the random order of the items changed from day to day (Section~\ref{ssec:models}). One more day of news can hardly change the judgment of actions taken months or years earlier, and the jump appears for every model, so it most likely comes from the change of the order: the same-day difference shows the randomness of a chatbot given an identical prompt, and a different order of the items adds to it. The jump depends little on how far an item moved in the list (somewhat more for Gemini), so it reflects the change of the prompt as a whole rather than a preference for early or late positions.
After the first day, the drift grows slowly: from $d=1$ to $d=25$, by 0.06 to 0.14 points for the actions, and by less, or not at all, for the reforms. The exception is Mistral, whose drift for the actions keeps rising, from 0.67 to about 1.0, mostly because of a single change of its system in the middle of the month (Section~\ref{sssec:model_changes_mistral}). 

For the actions, a model's responses 25 days apart remain closer to each other than to another model's responses on the same day: after 25 days, the drift of Gemini, Sonnet and the two Opus settings is 0.46 to 0.62, while their same-day differences from the other models are 0.67 to 0.88 (Figure~\ref{fig:MM_diff_scores}). Waiting a month thus changes a judgment less than switching to another chatbot. The exceptions are the two Opus settings, which differ from each other by only 0.38, and Mistral, whose drift reaches 1.0 because of its change of system.

The drift is not the same for all items. For each model, Figures~\ref{fig:drift_model_M}--\ref{fig:drift_model_Om} in Appendix~\ref{app:drift_by_model} show the median over the items and the range between their first and third quartiles. The median is somewhat below the mean, which is raised by a minority of items that drift more: after 25 days, a quarter of the actions drift by more than 0.6 to 0.7 points (1.15 for Mistral).

\subsection{Changes in the Models}\label{ssec:model_changes}

Figure~\ref{fig:daily_means} (Appendix~\ref{app:model_changes}) shows the daily means of the scores and of the other fields, averaged over the items and the two runs of each day. Besides the day-to-day noise, it shows changes that last for several days. At least three of them coincide with identifiable events and can be confirmed statistically:
\begin{enumerate}[topsep=0pt,itemsep=-1ex,partopsep=1ex,parsep=1ex]
    \item Claude Sonnet 5: on several days it reported its knowledge cutoff as August 2026 instead of January 2026, and its responses changed on exactly those days (Section~\ref{sssec:model_changes_sonnet}).
    \item Mistral: its chat interface changed in the middle of the month, and its responses changed with it for the rest of the month (Section~\ref{sssec:model_changes_mistral}).
    \item Claude Opus 5: on five days in the week before the release of Opus~5.5, both effort settings responded markedly more neutrally (Section~\ref{sssec:model_changes_opus}).
\end{enumerate}

\begin{table}[t]
\centering
\small
\begin{tabular}{@{}lrrr@{}}
\toprule
 & August & January & $p$ \\
\midrule
Comment length & 82.5 & 71.0 & $<$0.0001 \\
Reforms score & 3.41 & 3.67 & 0.0004 \\
Reform contestedness & 3.49 & 3.78 & $<$0.0001 \\
Share of action score${}=0$ & 4.6\% & 2.1\% & 0.0203 \\
\bottomrule
\end{tabular}
\caption{Sonnet's answers on the days it reported a different knowledge cutoff. August: the 6 days (6--9, 11, 12) on which Sonnet, asked daily for its knowledge cutoff, answered ``End of August 2026''; January: the 23 days on which it answered ``January 2026''. Day 16, when it gave both answers, is left out. Each quantity is averaged per day over items and both runs, then over the days of each group. Comment length is in characters and pools the comments on actions and reforms; reforms score is the magnitude (1--5). $p$: exact two-sided permutation test on the daily values, over all 475{,}020 ways of choosing 6 of the 29 days.}
\label{tab:sonnet-cutoff}
\end{table}

\begin{figure}[t]
    \centering
    \includegraphics[width=\columnwidth]{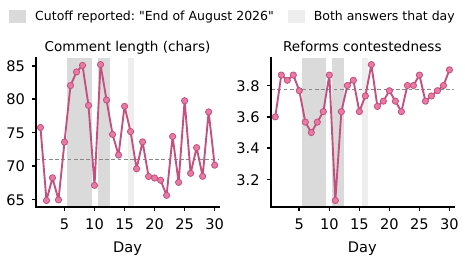}
    \caption{Sonnet's daily mean comment length (left) and mean contestedness of the reforms (right). Shaded: the days on which Sonnet reported its knowledge cutoff as ``End of August 2026''; lighter: day 16, when it gave both responses. Dashed: the mean over the days on which it reported ``January 2026''.}
    \label{fig:sonnet_cutoff_days}
\end{figure}

\subsubsection{Sonnet 5}\label{sssec:model_changes_sonnet}
Each day we also asked the Claude models for their knowledge cutoff and the current date. Sonnet~5 usually answered ``January 2026'' or ``End of January 2026'', but on six days (18--21 and 23--24 September; days 6--9, 11 and 12) it answered ``End of August 2026''; on 28 September (day 16) it gave both answers at different times of the day. Table~\ref{tab:sonnet-cutoff} shows that its responses to our prompts changed on exactly those days. Its comments were about 16\% longer, and it rated the reforms both as having smaller effects and as less contested; it also gave a score of zero to actions about twice as often. Its scores of the actions, by contrast, changed little: neither their size nor its confidence differed significantly between the two groups of days. The self-reported cutoff is not a reliable identifier of a model, but here it varies together with how the model writes and judges, which suggests that a different model or configuration was served under the same name on those days. It was not the released Sonnet~5.5: when that model appeared in the interface on 28 September, it reported a cutoff of June 2026.

Figure~\ref{fig:sonnet_cutoff_days} shows two of these quantities day by day. Comment length varies from day to day on its own (it is also high on days 15, 25 and 29), so the evidence lies in the clustering of high values on the shaded days rather than in any single day. The clearest sign is day 10, the one day inside this period on which Sonnet~5 again reported ``January 2026'': both comment length and contestedness return to their usual level on that day, and depart from it again on day 11. A user of the chat interface had no way to notice this change except by asking the model about itself.

\subsubsection{Mistral}\label{sssec:model_changes_mistral}

\begin{table}[t]
\centering
\small
\begin{tabular}{@{}lrrr@{\,}l@{}}
\toprule
 & Before & After & $p$ & \\
\midrule
Comment length, actions & 30.8 & 89.5 & $<$0.0001 & $^\dagger$ \\
Comment length, reforms & 22.6 & 86.5 & $<$0.0001 & $^\dagger$ \\
Action $|$score$|$ & 3.20 & 2.50 & $<$0.0001 &  \\
Action confidence & 4.05 & 2.92 & $<$0.0001 & $^\dagger$ \\
Reforms score & 3.80 & 3.16 & $<$0.0001 &  \\
Reforms confidence & 3.74 & 2.72 & $<$0.0001 & $^\dagger$ \\
Run-to-run difference & 0.66 & 0.46 & 0.0031 &  \\
Distance to other models & 0.85 & 0.51 & $<$0.0001 & $^\dagger$ \\
\bottomrule
\end{tabular}
\caption{Mistral before (13--26 September, days 1--14) and after (27 September--12 October, days 15--30) its switch to the new Vibe experience. Each quantity is computed per day over items and both runs, then averaged over the days of each period. Comment length is in characters. Reforms score is the magnitude (1--5). Run-to-run difference: mean absolute difference between the two daily runs' action scores (Eq.~\ref{eq:diff_MM_diag} for Mistral alone). Distance to other models: mean absolute difference between Mistral's action score and the mean score of the other four models on the same day. $p$: two-sided permutation test on the daily values, 10$^6$ random splits of the 30 days. $^\dagger$Every day after the switch lies beyond every day before it.}
\label{tab:mistral-switch}
\end{table}

On 27 September (day 15) the Mistral interface we used changed: the incognito mode was renamed ``Temporary chat'', and on the following day the separate Think option disappeared. This was our account's migration to Mistral's new experience, which merged the earlier Chat and Work interfaces; according to the provider's release notes, free and paid accounts were migrated in phases starting 14 September \citep{mistral2026releasenotes}. Until then our Mistral condition used the Think mode of the legacy Chat, documented as powered by Mistral's Magistral reasoning models \citep{mistral2026thinkmode}. In the new experience, Think is a toggle that switches to ``deeper, agentic workflows'', and the model behind it is not stated \citep{mistral2026vibe}. Our prompts did not change.

Table~\ref{tab:mistral-switch} compares Mistral's responses in the 14 days before and the 16 days after the switch. All eight quantities changed. The comments became three to four times longer, confidence fell by about one point on both actions and reforms, and the scores became milder. The new system was also more consistent: its two daily runs differed less from each other, and its action scores moved much closer to the mean of the other four models, a distance that fell by 40\%. For the reforms, that distance did not change (0.75 in both periods). For five of the eight quantities the two periods do not overlap at all: every day after the switch lies beyond every day before it.

Figure~\ref{fig:daily_means} shows that the change was a single step rather than a gradual drift. Mistral's daily confidence and scores fall on day 15 and stay at the new level for the rest of the month. The steady rise of Mistral's drift curves in Figure~\ref{fig:drift_models} therefore reflects mostly this one step, and any analysis that assumes a fixed system should treat the Mistral data as coming from two systems, days 1--14 and days 15--30. A user asking the same chatbot the same question would have received systematically different, and more cautious, judgments after a change that showed up in the interface mainly as a renamed option.

\subsubsection{Opus 5}\label{sssec:model_changes_opus}

\begin{table}[t]
\centering
\small
\setlength{\tabcolsep}{3.5pt}
\begin{tabular}{@{}lrrrr@{}}
\toprule
 & \multicolumn{2}{c}{Before 22 Sep} & From & \\
\cmidrule(lr){2-3}
 & Unusual & Other & 22 Sep & $p$ \\
\midrule
\multicolumn{5}{@{}l}{\emph{Opus high}} \\
\quad Action $|$score$|$ & 2.04 & 2.80 & 2.81 & 0.0004 \\
\quad Action confidence & 2.68 & 3.16 & 3.18 & 0.0005 \\
\quad Share of score${}=0$ & 8.0\% & 0.3\% & 1.0\% & 0.0019 \\
\quad Hedging comments & 12.5\% & 6.2\% & 7.0\% & 0.0242 \\
\quad Reforms score & 3.47 & 3.80 & 3.77 & 0.0553 \\
\quad Reforms confidence & 2.29 & 2.88 & 2.85 & 0.0185 \\
\midrule
\multicolumn{5}{@{}l}{\emph{Opus max}} \\
\quad Action $|$score$|$ & 1.97 & 2.89 & 2.86 & 0.0007 \\
\quad Action confidence & 2.56 & 3.00 & 3.06 & $<$0.0001 \\
\quad Share of score${}=0$ & 14.3\% & 2.5\% & 0.2\% & $<$0.0001 \\
\quad Hedging comments & 12.5\% & 6.2\% & 4.8\% & 0.0005 \\
\quad Reforms score & 3.08 & 3.85 & 3.86 & 0.0005 \\
\quad Reforms confidence & 2.19 & 2.67 & 2.69 & 0.0004 \\
\bottomrule
\end{tabular}
\caption{Opus~5 in the week before Opus~5.5 was released. Each quantity is computed per day over items and both runs, then averaged over three groups of days: Unusual, the days 3, 4, 6, 7 and 9 (15--21 September); Other, the other days before the release (1, 2, 5, 8); and the days from the release of Opus~5.5 on 22 September on (10--30). Share of score${}=0$ and hedging comments refer to the actions; a comment counts as hedging if it contains ``uncertain'', ``unclear'', ``contested'', ``disputed'', ``debat(e)'', ``mixed'', ``too recent'' or ``too early''. Reforms score is the magnitude (1--5). The unusual days were identified in the data and are shown for description; $p$ tests the split fixed by the release date, all days before it (1--9) against all days from it (10--30): two-sided permutation test on the daily values, 10$^6$ random splits.}
\label{tab:opus-dip}
\end{table}

On five of the seven days from 15 to 21 September (days 3, 4, 6, 7 and 9), both Opus settings responded markedly more neutrally (Table~\ref{tab:opus-dip}). Their scores moved toward zero for all four administrations. The share of zero scores (``no material long-run effect''), otherwise between 0.2\% and 2.5\%, rose to 8\% for Opus high and 14\% for Opus max, mostly for actions of the first Trump and the Biden administrations; comments containing words of doubt such as ``uncertain'' or ``contested'' became about twice as frequent; and confidence fell by about half a point on both actions and reforms. Other signs stayed the same: the comments were as long as on other days, and asked for its knowledge cutoff, Opus answered ``end of May 2026'' every day of the experiment. The two settings moved together (the correlation of their daily mean absolute score over the 30 days is 0.93), so the cause lay in what was served under both rather than in the effort setting.

The episode ended on 22 September (day 10), the day Opus~5.5 was released \citep{anthropic2026opus55} and appeared in the model selector. We continued to use Opus~5, which from that day responded at its usual level for the rest of the month. Anthropic's published system prompt for Opus~5 shows a single version, dated 24 July 2026 \citep{anthropic2026opus5prompts}, so no documented change to it falls within the experiment, and we do not know what caused the episode. Because the five days were identified in the data, Table~\ref{tab:opus-dip} reports $p$-values for a split fixed by the release date instead, all days before it against all days from it. Even this diluted comparison, which counts four ordinary days in the first group, is significant at the 5\% level for 11 of the 12 quantities. Unlike Sonnet's change, this one left no trace in the model's responses about itself or in the style of its comments. It was visible only in its judgments.

\subsection{Reasoning}\label{sec:Reasoning}
\subsubsection{Leading Dimension of the Reforms}\label{sssec:reform_dimension}
The models agreed almost completely on which dimension determined the magnitude of a reform. All the models, on almost every day and in almost every run, assigned the reforms I01--I06 (political institutions and state capacity) to \textit{institutional}, the reforms I07--I14 to \textit{income}, and I15 (insurance against catastrophes) to \textit{lives}.

Only 51 of the 4{,}500 responses (1.1\%) chose a different dimension: 22 of Mistral, 18 of Opus high, 5 of Gemini, 4 of Sonnet and 2 of Opus max. The departures concentrate on the reforms whose effects plausibly fall on more than one dimension:
\begin{enumerate}[topsep=0pt,itemsep=-1ex,partopsep=1ex,parsep=1ex]
    \item Early childhood (I11) as \textit{lives} instead of \textit{income}: 17.
    \item State capacity (I06) as \textit{income} instead of \textit{institutional}: 14.
    \item Catastrophe insurance (I15) as \textit{income} instead of \textit{lives}: 7.
    \item Fiscal path (I14) as \textit{institutional}: 5.
    \item Reading and math (I12) as \textit{lives}: 5.
\end{enumerate}
The departures show no pattern over time: they are scattered across the month, and Mistral's switch on day 15 made no difference (9 departures before it, 13 after).

\subsubsection{Consistency of Comments}\label{sssec:comments_consistency}

As requested by the prompts (Figures~\ref{fig:p2-prompt} and~\ref{fig:f-prompt}), each response contains a short comment (11 words in the median) giving the reason for the score. To ask whether two models give the same kinds of reasons, we measure how well the comments of one model on an item identify the comments of the other model on the same item among all its comments. This lets us use standard retrieval measures to evaluate the consistency of the reasons.

\begin{figure*}[t]
  \centering
  \includegraphics[width=\textwidth]{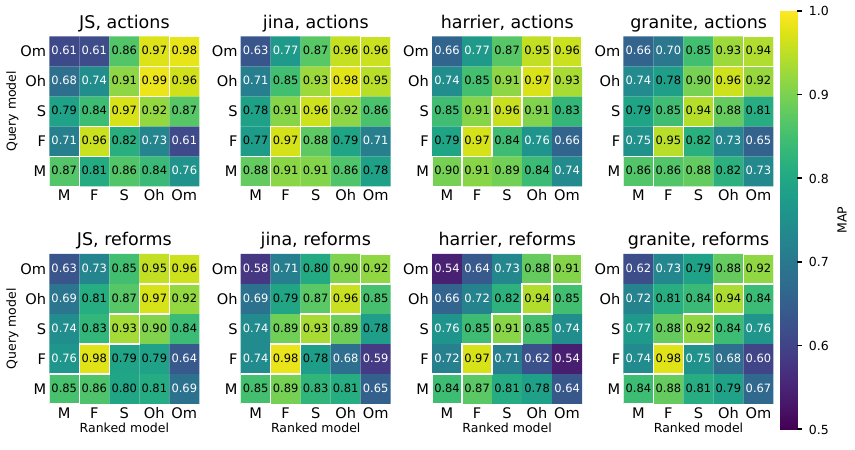}
  \caption{Retrieval of comments across models: mean average precision for
  each query model (rows) and ranked model (columns), with Jensen--Shannon
  (JS) and three embedding models, for the actions (top) and the reforms
  (bottom). Diagonal cells (outlined): query and ranked comments from
  different runs of the same model. Chance level is about 0.025 for the
  actions and 0.067 for the reforms.}
  \label{fig:ret-map}
\end{figure*}

Let $c(M,x,t,i)$ be the comment of model $M$ on item $x$ on day $t$ in run $i$, and $C(M,x,i)$ the set of its 30 comments on $x$ in run $i$:
\begin{equation}\label{eq:set_of_comments}
 C(M,x,i)=\{c(M,x,t,i): t\in T\}
\end{equation}
For the similarity $S(c,C)$ between a single comment $c$ and a set of comments $C$, we use the Jensen--Shannon divergence or embedding models.

With the Jensen--Shannon divergence:
\begin{equation}\label{eq:sim_js}
 S_{JS}(c,C)=1-\mathrm{JS}\bigl(p(c),\ \tfrac{1}{|C|}\sum_{c'\in C} p(c')\bigr)
\end{equation}
where $p(c)$ is the distribution of the content words of $c$ (lower-cased words, without a short list of function words and without stemming), and $\mathrm{JS}$ is the Jensen--Shannon divergence in bits \citep{lin1991divergence}, which has long been used to compare the word distributions of texts, for example in the evaluation of summaries without human references \citep{louis-nenkova-2009-automatically}. Each comment in $C$ thus has equal weight, whatever its length.

With an embedding model $e$:
\begin{equation}\label{eq:sim_emb}
 S_{e}(c,C)=\cos\bigl(e(c),\ \tfrac{1}{|C|}\sum_{c'\in C} e(c')\bigr)
\end{equation}
where $e(c)$ is the unit-length embedding of $c$.

We use three embedding models of different architectures: 
\begin{enumerate}[topsep=0pt,itemsep=-1ex,partopsep=1ex,parsep=1ex]
    \item \textit{gina}: jina-embeddings-v5-text-small, a decoder built on Qwen3 \citep{jina2026v5small}.
    \item \textit{harrier}: harrier-oss-v1-270m, a decoder built on Gemma~3 \citep{harrier2026}.
    \item \textit{granite}: granite-embedding-97m-multilingual-r2, a ModernBERT encoder \citep{granite2026r2}.
\end{enumerate}
Each model is used with the prompt or task setting that its model card specifies for text similarity.

For a query model $M_1$, an item $x$ and a ranked model $M_2$, all $30|X|$ comments of $M_2$ in run~2 are ranked by $S\bigl(c(M_2,x',t,2),C(M_1,x,1)\bigr)$, and the 30 comments with $x'=x$ are relevant. We summarize each ranking by its average precision and by its R-precision, the share of relevant comments among the first 30 \citep{manning2008ir}, and average over the $|X|$ items and over both assignments of the runs (query from run~1 and ranked comments from run~2, and the reverse). This gives one value for each of the 25 ordered pairs $(M_1,M_2)$. On the diagonal, the query and the ranked comments come from different runs of the same model, so the diagonal is a run-to-run baseline. A random ranking gives about $1/|X|$: 0.025 for the actions and 0.067 for the reforms.

Figure~\ref{fig:ret-map} shows the mean average precision (MAP). The comments are highly specific to the items: a model's own comments from the other run are retrieved with MAP between 0.92 and 0.95 on average, depending on the similarity measure, and another model's comments with 0.80--0.84 for the actions and 0.74--0.80 for the reforms, far above chance. Part of this success comes from topic words that every model uses for a given item, so the differences between pairs reflect how similarly the models phrase their reasons. The Claude models form a family: the two Opus settings are nearly interchangeable (0.92--0.97 for the actions, 0.84--0.95 for the reforms), and Sonnet is close to both. Gemini and Mistral each comment in their own way. The weakest pairs, 0.54--0.66, always join Opus max with Mistral or Gemini. R-precision gives the same picture (Figure~\ref{fig:ret-rprec} in Appendix~\ref{app:comments_consistency}).

These conclusions do not depend on the similarity measure. Across the four measures, the off-diagonal cells are ordered alike: Spearman correlations between measures range from 0.84 to 0.98, and from 0.93 to 0.98 among the three embedding models. Nor do they depend on the ranking metric (Table~\ref{tab:ret-metrics}). R-precision, MAP and nDCG truncated at $K=30$, and full nDCG order the cells as MAP does ($\rho \geq 0.98$). The reciprocal rank of the first relevant comment (MRR) is at its ceiling: the top-ranked comment is about the right item in almost every query, and its weak correlation with MAP reflects noise near 1. AUC follows MAP within a narrow range: a comment on the right item ranks above a comment on another item in 86.5\% to 99.9\% of all such pairs, for every pair of models.

\begin{table}[t]
\centering
\small
\setlength{\tabcolsep}{4pt}
\begin{tabular}{@{}lcccr@{}}
\toprule
Metric & $\rho$ & $\tau_b$ & Range & $\geq 0.99$ \\
\midrule
MAP & & & 0.54--0.97 & 0\% \\
R-precision & 0.98--1.00 & 0.88--0.98 & 0.49--0.93 & 0\% \\
MAP@$K$ & 0.98--1.00 & 0.89--1.00 & 0.40--0.93 & 0\% \\
nDCG@$K$ & 0.99--1.00 & 0.92--0.98 & 0.57--0.95 & 0\% \\
nDCG & 0.98--0.99 & 0.91--0.97 & 0.83--0.99 & 2\% \\
AUC & 0.93--0.97 & 0.79--0.89 & 0.86--1.00 & 22\% \\
MRR & 0.29--0.73 & 0.23--0.61 & 0.81--1.00 & 54\% \\
\bottomrule
\end{tabular}
\caption{Agreement of other ranking metrics with MAP. For each similarity measure, and separately for the actions and the reforms, each metric gives a $5\times 5$ matrix with one cell for each pair of a query model and a ranked model, as in Figure~\ref{fig:ret-map}. Its 20 off-diagonal cells (pairs of different models) are compared with those of MAP, and the ranges are over the 8 matrices (4 similarity measures, for the actions and for the reforms). $\rho$: Spearman rank correlation with MAP; $\tau_b$: Kendall's tau-b with MAP, which accounts for ties (frequent for MRR). Range: smallest and largest off-diagonal value over all matrices; $\geq 0.99$: share of off-diagonal cells at 0.99 or above.}
\label{tab:ret-metrics}
\end{table}

\begin{table}[t]
\centering
\small
\setlength{\tabcolsep}{4pt}
\begin{tabular}{@{}lcccccc@{}}
\toprule
 & \multicolumn{3}{c}{Actions} & \multicolumn{3}{c}{Reforms} \\
\cmidrule(lr){2-4}\cmidrule(l){5-7}
Model & Own & Query & Ranked & Own & Query & Ranked \\
\midrule
M & 0.88 & 0.84 & 0.73 & 0.85 & 0.79 & 0.69 \\
F & 0.96 & 0.75 & 0.82 & 0.98 & 0.70 & 0.81 \\
S & 0.96 & 0.86 & 0.88 & 0.92 & 0.82 & 0.80 \\
Oh & 0.97 & 0.84 & 0.86 & 0.95 & 0.80 & 0.82 \\
Om & 0.96 & 0.79 & 0.80 & 0.93 & 0.75 & 0.72 \\
\bottomrule
\end{tabular}
\caption{Each model as query and as ranked model: MAP averaged over the 4 similarity measures. Own: the diagonal cell, the model's comments retrieved by its own query from the other run. Query: mean over the other four models of the model's row, i.e.\ how well its averaged comments retrieve their comments. Ranked: mean of the model's column, i.e.\ how well its comments are retrieved by the other models' queries.}
\label{tab:ret-asym}
\end{table}

The matrices are not symmetric, and the difference is informative (Table~\ref{tab:ret-asym}). Mistral's averaged comments retrieve other models' comments well (0.84 for the actions), but its own comments are the hardest to retrieve (0.73), and they are also the least consistent between its own two runs (0.88, against 0.96--0.97 for the other models). This fits the change of system on 27 September (Section~\ref{sssec:model_changes_mistral}), which mixes two styles of comments in Mistral's data. Gemini shows the opposite pattern: its comments are the most consistent of all (0.96 for the actions, 0.98 for the reforms) and are found easily by other models' queries (0.82), but its own averaged comment retrieves other models' comments least well (0.75 and 0.70): Gemini phrases its reasons in a way of its own.

Word overlap and embeddings differ in one respect. With Jensen--Shannon, comments on reforms are retrieved across models as well as comments on actions (off-diagonal MAP 0.80 for both). With the embedding models they are retrieved less well (0.74--0.77 against 0.80--0.84). A plausible explanation is that the reforms are argued in a shared vocabulary of growth, institutions and long-run costs. Embeddings see comments on different reforms as close in meaning and confuse them more often, whereas word overlap remains anchored in the specific terms of each reform.

\section{Conclusion}\label{sec:Conclusion}
We asked five chatbot configurations, through their consumer chat interfaces, to judge 40 presidential actions and 15 hypothetical reforms twice a day for 30 days. Our experiment suggests that chatbots can give robust judgments of both real policies and hypothetical reforms: the scores, the characteristics that accompany them, and the comments that explain them distinguish the items well, and are reasonably stable against the choice of chatbot, the drift from day to day, and even the changes in the chatbot services. In terms of the questions of the introduction:
\begin{enumerate}[topsep=0pt,itemsep=-1ex,partopsep=1ex,parsep=1ex]
    \item Choice of chatbot. The models agree on the sign of each administration's mean score and on the ranking of the administrations, and two models asked about the same action on the same day give opposite signs in about 4\% of the cases. The models differ more from each other than from themselves, and the three Claude configurations are closer to each other than to Gemini or Mistral, in their scores, in their other fields, and in the wording of their comments.
    \item Drift from day to day. Responses given on different days differ more than the two responses of the same day, most likely because the order of the items changed from day to day while it was the same for both runs of a day. After this first step, the difference grows slowly over the month. For the actions, a model's responses given 25 days apart still differ less than its responses and those of another provider's model given on the same day; the exception is Mistral, whose system changed in the middle of the month.
    \item Variation within a day. Two runs of the same model on the same day differ by 0.27 to 0.55 points on the action scale, and give opposite signs in less than 1\% of the cases for the Claude models and in almost 4\% for Gemini. This is the randomness of a chatbot given an identical prompt; a different order of the items, as between consecutive days, adds 0.08 to 0.20 points.
    \item Real and hypothetical decisions. The judgments of the hypothetical reforms behave much like those of the real actions: the responses cluster as closely around each item's median, the models differ in the same pattern, and the drift over the month is, if anything, smaller. The reforms are, however, judged with lower confidence, about as low as the most recent actions, and the comments on different reforms are harder to tell apart by meaning than by wording.
\end{enumerate}

The three changes that we identified show the main risk for a user: the system behind a chatbot's name can change without notice, and shift its judgments for days or for good, while the interface shows little or nothing of it. A judgment that matters is therefore best checked with more than one chatbot, and on more than one day. The dataset that we release allows further analysis of these judgments, including the analysis of individual items, which we have left aside.

\section*{Limitations}\label{sec:Limitations}
Our experiment and analysis have several limitations:
\begin{enumerate}[topsep=0pt,itemsep=-1ex,partopsep=1ex,parsep=1ex]
    \item The interval of 30 days, although interesting because it falls in the heated weeks before the midterm elections, is short compared with the time scales of the presidential actions and the hypothetical reforms that we consider.
    \item The set of models could be wider: we used five configurations from three providers, three of them from Anthropic.
    \item Each set of items was judged with a single prompt, in English. We did not test how the judgments depend on the wording, format or language of the prompt, which can affect the responses of LLMs to political questions \citep{rottger-etal-2024-political,moore-etal-2024-large}, nor whether naming the administration of each action influences its score.
    \item Both runs of a day used the same order of the items, and the order changed every day. The same-day difference therefore leaves out the effect of the order, and the design cannot fully separate the effect of the order from that of the passing one day.
    \item Both the actions and the reforms were selected with the help of Claude Opus 5, one of the models that we tested, and the selection may reflect its view of what is consequential.
    \item Robustness is not correctness. Agreement between the models and stability over time do not show that the judgments are right; models trained on similar data may share the same errors. There is no reference response against which the judgments could be checked.
    \item Chat interfaces give the user neither control over nor a record of the model version, system prompt and tools behind an response. The three changes that we describe were identified after the fact, from the data and from the models' responses about themselves; smaller changes may have gone unnoticed.
    \item We left out the analysis of the political or economic reasons for the scores themselves (Figures~\ref{fig:p2_admin_ranges} and~\ref{fig:item_dots_all}).
    \item We did not analyze the individual items in depth: why particular actions or reforms are judged differently by different models, or how their judgments change over time.
    \item The 50-year horizon of the prompts is a guess at a trade-off between a bias toward short-term effects and the many unknowns of the distant future. Some discounting of later effects might be more appropriate, especially because large and unforeseeable changes are likely within such a horizon \citep{bostrom2019vulnerable}.
\end{enumerate}

\bibliography{custom}

\appendix

\onecolumn
\section{Presidential Actions and Hypothetical Reforms}\label{app:items}
\subsection{40 Presidential Actions}\label{sapp:items_actions}
Table~\ref{tab:p2-actions} lists the 40 presidential actions, and Figure~\ref{fig:p2-prompt} shows the prompt with which they were judged.
\begingroup
\small
\setlength{\tabcolsep}{5pt}
\setlength{\LTcapwidth}{\linewidth}
\ifdefined\captionsetup\captionsetup{position=below}\fi
\begin{xltabular}{\linewidth}{@{}l c c >{\raggedright\arraybackslash}X@{}}
\toprule
Id & Year & Admin & Action and description \\
\midrule
\endfirsthead
\bottomrule
\caption{The 40 presidential actions, in chronological order. Year: when the action was launched; actions spanning several years are listed by their first year. Admin: O~=~Obama, T1~=~Trump (first term), B~=~Biden, T2~=~Trump (second term). Names and descriptions are as given to the models; the items were shuffled into a new order each day, the same for both runs of the day.}\label{tab:p2-actions}\\
\endlastfoot
\multicolumn{4}{@{}l}{\itshape Table~\thetable{} (continued)}\\
\toprule
Id & Year & Admin & Action and description \\
\midrule
\endhead
\midrule
\multicolumn{4}{@{}r@{}}{\itshape Table~\thetable{} continued on next page}\\
\endfoot
I21 & 2009 & O & \textbf{American Recovery and Reinvestment Act.} $\sim$\$800B in tax cuts, state aid, infrastructure, and the first large federal clean-energy push. \\
I13 & 2009 & O & \textbf{Auto rescue and bank stress tests.} Managed bankruptcy of GM and Chrysler plus forced recapitalization of the largest banks. \\
I02 & 2010 & O & \textbf{Affordable Care Act.} Health insurance law expanding Medicaid eligibility and creating subsidized individual exchanges. \\
I25 & 2010 & O & \textbf{New START treaty.} Ratified caps on deployed US and Russian strategic warheads with mutual inspections. \\
I29 & 2010 & O & \textbf{Dodd-Frank Act and the CFPB.} Bank capital and derivatives rules plus a new federal consumer financial regulator. \\
I37 & 2011 & O & \textbf{Mercury and Air Toxics Standards.} EPA limits on coal plant emissions credited with thousands of avoided deaths each year. \\
I17 & 2012 & O & \textbf{DACA.} Deferred deportation and work permits for roughly 800,000 people brought to the US as children. \\
I05 & 2012 & O & \textbf{2012 CAFE fuel economy standards.} Doubled required fleet efficiency by 2025, cutting oil demand and vehicle emissions. \\
I33 & 2015 & O & \textbf{Iran nuclear deal (JCPOA).} Multilateral agreement freezing Iran's nuclear program in exchange for sanctions relief. \\
I09 & 2016 & O & \textbf{Joining the Paris Agreement.} US entry into the global climate framework with a pledge to cut emissions 26--28\% by 2025. \\
\midrule
I20 & 2017 & T1 & \textbf{Reshaping the federal judiciary.} Three Supreme Court seats and 230+ lifetime judgeships, shifting constitutional law for decades. \\
I28 & 2017 & T1 & \textbf{Environmental deregulation agenda.} Repeal of the Clean Power Plan, methane limits, and federal clean-water protections. \\
I07 & 2017 & T1 & \textbf{Tax Cuts and Jobs Act.} Permanent corporate rate cut and temporary individual cuts, adding trillions to federal debt. \\
I24 & 2018 & T1 & \textbf{Withdrawal from the Iran nuclear deal.} Exit from JCPOA and maximum-pressure sanctions, ending inspection limits on Iran's program. \\
I16 & 2018 & T1 & \textbf{China trade war tariffs.} Section 301 duties on $\sim$\$350B of imports, breaking the postwar bipartisan free-trade consensus. \\
I11 & 2020 & T1 & \textbf{Federal COVID-19 pandemic response.} Downplaying of the virus and deferral of testing and mitigation to states through 2020. \\
I36 & 2020 & T1 & \textbf{Doha agreement with the Taliban.} Deal setting a US withdrawal date that excluded the sitting Afghan government from talks. \\
I32 & 2020 & T1 & \textbf{CARES Act.} \$2.2T pandemic relief with expanded unemployment, direct payments, and small business loans. \\
I03 & 2020 & T1 & \textbf{Operation Warp Speed.} Program that funded and pre-purchased COVID-19 vaccines at risk during development. \\
\midrule
I19 & 2021 & B & \textbf{National COVID-19 vaccination campaign.} Free mass distribution of vaccines credited with several hundred thousand US lives saved. \\
I12 & 2021 & B & \textbf{American Rescue Plan Act.} \$1.9T relief that sharply cut child poverty and contributed to the 2021--22 inflation surge. \\
I23 & 2021 & B & \textbf{Withdrawal from Afghanistan.} Completed the troop exit, ending the 20-year war amid collapse of the Afghan state. \\
I35 & 2021 & B & \textbf{Expanded parole and asylum admissions.} Humanitarian entry programs amid record crossings, admitting millions and reshaping politics. \\
I27 & 2021 & B & \textbf{Infrastructure Investment and Jobs Act.} \$1.2T for roads, bridges, transit, water lines, grid, and rural broadband over a decade. \\
I04 & 2022 & B & \textbf{Arming Ukraine and enlarging NATO.} Roughly \$175B in aid and accession of Finland and Sweden after Russia's full-scale invasion. \\
I08 & 2022 & B & \textbf{Inflation Reduction Act.} Clean-energy tax credits plus first Medicare authority to negotiate prescription drug prices. \\
I31 & 2022 & B & \textbf{CHIPS and Science Act.} \$52B in subsidies to relocate advanced semiconductor fabrication onto US soil. \\
I15 & 2022 & B & \textbf{Semiconductor export controls on China.} Rules blocking advanced chips and tooling, defining the long-run US--China technology rivalry. \\
\midrule
I01 & 2025 & T2 & \textbf{Mass deportation campaign.} Large ICE buildup, interior raids, and revocation of protected status for millions of residents. \\
I18 & 2025 & T2 & \textbf{Exit from WHO and other institutions.} Withdrawal from the WHO, Paris Agreement, and dozens of other international bodies. \\
I34 & 2025 & T2 & \textbf{Pardons of January 6 offenders.} Clemency for roughly 1,500 convicted over the Capitol attack, including violent offenders. \\
I22 & 2025 & T2 & \textbf{Cuts to federal research funding.} Deep reductions to NIH and NSF grants and to university indirect-cost recovery. \\
I26 & 2025 & T2 & \textbf{Impoundment and federal workforce purge.} Withholding of appropriated funds and mass firings across the career civil service. \\
I40 & 2025 & T2 & \textbf{Dismantling of USAID.} Shutdown of the main foreign aid agency and cancellation of most global health programs. \\
I30 & 2025 & T2 & \textbf{Universal tariff regime.} Across-the-board import duties, voided under IEEPA in 2026 and reimposed via other statutes. \\
I38 & 2025 & T2 & \textbf{Overhaul of federal vaccine policy.} Replacement of ACIP advisers, narrowed recommendations, and cancelled mRNA vaccine contracts. \\
I39 & 2025 & T2 & \textbf{Domestic deployment of the military.} National Guard and federal troops sent into US cities over the objections of local officials. \\
I14 & 2025 & T2 & \textbf{One Big Beautiful Bill Act.} Made 2017 tax cuts permanent while cutting Medicaid and SNAP and repealing clean-energy credits. \\
I06 & 2025 & T2 & \textbf{Pressure on Federal Reserve independence.} Attempted removal of a sitting governor and sustained public demands for rate cuts. \\
I10 & 2026 & T2 & \textbf{Military intervention in Venezuela.} Strikes on Caracas and capture of Nicolas Maduro without congressional authorization. \\
\end{xltabular}
\endgroup

\ifdefined\FVExtraRobustCommand\else\PackageError{prompt figure}{Add \string\usepackage{fvextra} to the preamble}{The verbatim prompt box needs the fvextra package.} \fi
\begin{figure*}[tbp]
\begin{Verbatim}[fontsize=\footnotesize, frame=single, framesep=2mm, breaklines, breakafter=|, breakautoindent=false, breaksymbolindentleftnchars=2, breaksymbolsepleft=0.3em, breakindentnchars=7]
Score each presidential action below for its net effect on the long-term flourishing of the population of the United States, over the 50 years following the date that action was taken.

SCALE. The sign is the direction of the effect; the number is its magnitude.
+5 / -5  Foundational. Roughly 250,000 lives or more, about 2% permanent change in national income, or a change to a core institution that constrains all later policy. Not readily reversible.
+4 / -4  Major structural change. Roughly 50,000-250,000 lives, 1-2% of long-run income, or a nationwide effect reversible only at high cost.
+3 / -3  Large and durable. Roughly 10,000-50,000 lives, 0.3-1% of income, or tens of millions materially affected. Reversible at moderate cost.
+2 / -2  Moderate. A measurable effect on a defined sector or population. Reversible.
+1 / -1  Small, but directionally clear at national scale.
 0       No material long-run effect, or positive and negative effects that cancel.

CONFIDENCE. How sure you are of the score you assigned, including its sign.
5 very high, 4 high, 3 moderate, 2 low, 1 very low.

RULES.
Judge each action on its own merits, independently of the other actions listed and of any overall view of the administration that took it.
Score the action as taken, including consequences that became apparent later.
Base the score on the relevant empirical evidence as you understand it, not on a political judgment.
Assign a score to every item. Do not skip any.

OUTPUT. Output nothing except one line per item, in the order the items are listed below, in this exact form:
<id>|<score>|<confidence>|<comment>

score       integer from -5 to +5, always with an explicit + or - sign
confidence  integer from 1 to 5
comment     at most 100 characters, no pipe character

No preamble, no headings, no blank lines, no closing summary.

Each item is given as <id> | <administration> | <action name and description>

administration values: Obama, Biden, Trump1, Trump2. The last two are for the first and the second term of Trump.

ITEMS:
\end{Verbatim}
\caption{The prompt for the presidential actions, verbatim. In each run it was followed by the 40 items of Table~\ref{tab:p2-actions}, one per line, in a random order that changed daily and was the same for both runs of a day (e.g.\ \texttt{I17 | Obama | DACA \textemdash{} Deferred deportation and \ldots}). A $\hookrightarrow$ marks a line wrapped to fit the page; it is not a line break in the prompt.}\label{fig:p2-prompt}
\end{figure*}

\clearpage

\subsection{15 Hypothetical Reforms}\label{sapp:items_reforms}
Table~\ref{tab:f-actions} lists the 15 hypothetical reforms, and Figure~\ref{fig:f-prompt} shows the prompt with which they were judged.
\begingroup
\small
\setlength{\tabcolsep}{5pt}
\setlength{\LTcapwidth}{\linewidth}
\ifdefined\captionsetup\captionsetup{position=below}\fi
\begin{xltabular}{\linewidth}{@{}l >{\raggedright\arraybackslash}X@{}}
\toprule
Id & Action and description \\
\midrule
\endfirsthead
\bottomrule
\caption{The 15 hypothetical reforms. The models were asked to assume that each reform is taken now and fully supported by all branches of government. Names and descriptions are as given to the models; the items were shuffled into a new order each day, the same for both runs of the day.}\label{tab:f-actions}\\
\endlastfoot
\multicolumn{2}{@{}l}{\itshape Table~\thetable{} (continued)}\\
\toprule
Id & Action and description \\
\midrule
\endhead
\midrule
\multicolumn{2}{@{}r@{}}{\itshape Table~\thetable{} continued on next page}\\
\endfoot
I01 & \textbf{Expand the House.} Repeal the 1929 cap at 435 seats and set chamber size by the cube-root rule, roughly 690 seats, shrinking districts from $\sim$760,000 people to $\sim$480,000. Statutory, needs no amendment, and improves representation, committee capacity, and campaign economics at once. \\
I02 & \textbf{Reduce veto points.} Eliminate the legislative filibuster and prune the procedural chokepoints that let small groups block action. The defining trait of the US system is difficulty passing anything, which is also difficulty correcting mistakes. \\
I03 & \textbf{Abolish electoral college.} Replace it with a national popular vote by constitutional amendment, paired with uniform federal standards for registration, ballots, audits, and recounts. The election standards are not optional here: a national recount under 50 rulebooks would be worse than what exists now. \\
I04 & \textbf{Term-limit the judiciary.} Eighteen-year staggered Supreme Court terms with one appointment per presidential term. Removes the actuarial lottery that currently makes a single death worth more than most elections. \\
I05 & \textbf{Make the Constitution amendable.} Lower Article V thresholds and repeal the equal-suffrage entrenchment that permanently locks in the Senate's 68:1 representation gap between Wyoming and California. The US has the hardest constitution in the world to amend, which means structural errors compound indefinitely. \\
I06 & \textbf{Rebuild state capacity.} Civil service reform that raises pay and hiring speed for technical roles, strips procedural layering, and holds agencies to outcomes rather than process compliance. Every other item here depends on someone being able to execute it. \\
I07 & \textbf{Legalize abundant housing.} Preempt local zoning and discretionary review that block multifamily and infill construction, so supply tracks demand in high-wage metros. Restores geographic mobility and cuts the largest line item in most household budgets. \\
I08 & \textbf{Overhaul permitting.} Statutory deadlines for environmental review, narrow litigation windows, and federal siting authority for interstate transmission. Determines whether energy, transit, and housing take three years or fifteen. \\
I09 & \textbf{Rebuild legal immigration.} Expand skilled and family visas, untie green cards from single employers, clear the backlogs, and pair this with an enforcement and asylum system that functions. Offsets a shrinking working-age population. \\
I10 & \textbf{Scale basic research.} Sustained large increases to NIH, NSF, and DOE science, plus ARPA-style agencies, advance market commitments, and more funding aimed at young researchers. Among the highest measured long-run returns of any government spending. \\
I11 & \textbf{Invest in early childhood.} Universal prenatal and pediatric care, nutrition support, and high-quality childcare and pre-K, weighted toward low-income families. The returns show up decades later in earnings, health, and crime. \\
I12 & \textbf{Teach reading and math.} Tie federal education funding to structured literacy and evidence-based instruction, fund high-dosage tutoring, and raise teacher pay in exchange for higher selectivity. Human capital is the input to every other outcome. \\
I13 & \textbf{Fix healthcare prices.} Expand residency slots and scope of practice, allow reciprocal approval of drugs and devices cleared by peer regulators, and benchmark hospital and drug prices against reference rates. Targets the $\sim$17\% of GDP spent for outcomes peer countries match for far less. \\
I14 & \textbf{Stabilize the fiscal path.} Broaden the tax base toward land, consumption, and pollution while slowing age-related spending growth enough to hold debt-to-GDP steady. Keeps interest payments from crowding out everything else by the 2060s. \\
I15 & \textbf{Insure against catastrophes.} Standing pandemic capacity, serious biosecurity oversight, credible AI safety institutions, and continued nuclear risk reduction. Cheap insurance against events that would erase every other gain on this list. \\
\end{xltabular}
\endgroup

\ifdefined\FVExtraRobustCommand\else\PackageError{prompt figure}{Add \string\usepackage{fvextra} to the preamble}{The verbatim prompt box needs the fvextra package.} \fi
\begin{figure*}[tbp]
\begin{Verbatim}[fontsize=\footnotesize, frame=single, framesep=2mm, breaklines, breakafter=|, breakautoindent=false, breaksymbolindentleftnchars=2, breaksymbolsepleft=0.3em, breakindentnchars=0]
Score each hypothetical government action below for its net effect on
human flourishing in the United States over the next 50 years. Assume
the action is taken now and fully supported by all branches of
government.

Baseline for comparison: current law and current trends continuing for
50 years. Score in expectation, not modal outcome — a 10% chance of a
very large effect counts as roughly one tenth of that effect.

MAGNITUDE (1-5):
Use whichever dimension scores highest, and report which one in the
DIMENSION field. Common unit: permanent change in annual national
income as a share of GDP. For mortality, convert at roughly $10M per
premature death averted, spread over the 50 years. Rough conversions
are fine; the bands are wide.
5 Foundational. >=3% of GDP-equivalent, >=2M premature deaths averted,
  or restructures an institution that governs the majority of federal
  policymaking. Historical scale: Social Security and Medicare, the
  Interstate Highway System, universal public schooling.
4 Major. 1-3% of GDP-equivalent, 500k-2M lives, or restructures one
  significant institution. Historical scale: the Clean Air Act, the
  1965 Immigration Act, the GI Bill.
3 Large. 0.3-1% of GDP-equivalent, 150k-500k lives, or tens of millions
  materially affected. Historical scale: the ACA, the EITC.
2 Moderate. 0.1-0.3% of GDP-equivalent, 40k-150k lives, or a clear
  effect on one defined sector. Historical scale: the ADA, Dodd-Frank.
1 Modest or unclear. Below 0.1%, under 40k lives, effects that mostly
  cancel or effects too speculative or negative.

DIMENSION (lives / income / institutional):
Which dimension determined the magnitude score.

IRREVERSIBILITY (1-5):
How hard to undo within 50 years if it proves wrong. 5 = constitutional
or effectively permanent; 1 = repealable by ordinary legislation.

CONFIDENCE (1-5):
Your certainty in the magnitude. 5 = very high, 1 = very low.

CONTESTEDNESS (1-5):
How much informed experts disagree about the magnitude, independent of your
own confidence. 5 = deeply contested; 1 = not contested.

RULES:
Score each action on its own merits, assuming the other actions listed
do NOT happen.
Ignore implementation cost and political difficulty; score impact only.
Score every item. Do not skip any.

OUTPUT:
Output nothing except one line per item, in the order listed, in this
exact form:
<id>|<magnitude>|<dimension>|<irreversibility>|<confidence>|<contestedness>|<comment>
Comment is at most 100 characters and must not contain a pipe.
No preamble, no headings, no blank lines, no closing summary.

Each item below is given as <id> | <action name> | <action description>

ITEMS:
\end{Verbatim}
\caption{The prompt for the hypothetical reforms, verbatim. In each run it was followed by the 15 items of Table~\ref{tab:f-actions}, one per line, in a random order that changed daily and was the same for both runs of a day (e.g.\ \texttt{I01 | Expand the House | Repeal the 1929 cap \ldots}).}\label{fig:f-prompt}
\end{figure*}

\twocolumn


\section{Dataset}
\label{app:dataset}

All responses collected in the experiment are released as a dataset on
Hugging Face.\footnote{\url{https://huggingface.co/datasets/oleg-vasilyev/llm-policy-judgments}}
It consists of four files in JSON Lines format, each of which can be loaded
as a named configuration with the \texttt{datasets} library:
\begin{itemize}
\item \texttt{scores\_actions}: the 12{,}000 responses on the presidential
  actions, one row per response (30 days $\times$ 5 models $\times$ 2 runs
  $\times$ 40 actions);
\item \texttt{scores\_reforms}: the 4{,}500 responses on the hypothetical
  reforms (30 $\times$ 5 $\times$ 2 $\times$ 15);
\item \texttt{actions}: the 40 actions of Table~\ref{tab:p2-actions}, with
  the administration, name and description as given to the models, and the
  year in which each action was launched, which was not shown to the models;
\item \texttt{reforms}: the 15 reforms of Table~\ref{tab:f-actions}.
\end{itemize}

Each response row gives the day (1--30) and its date, the model and its symbol (M, F, S, Oh, Om), the run (1 or 2) and the
item id, followed by the fields requested in the prompt: score, confidence
and comment for the actions (Figure~\ref{fig:p2-prompt}); score,
dimension, irreversibility, contestedness, confidence and comment for the
reforms (Figure~\ref{fig:f-prompt}). The item ids of the actions and of the
reforms are separate: action I01 and reform I01 are different items.

The responses are given as parsed from the models' outputs. Comments longer
than the requested 100 characters are kept (914 of the 12{,}000 comments on
actions and 164 of the 4{,}500 on reforms), and the days affected by the
changes in the chat products (Tables~\ref{tab:sonnet-cutoff},
\ref{tab:mistral-switch} and~\ref{tab:opus-dip}) are included. The dataset
card describes every field.

\section{Scores of Individual Items, by Model}\label{app:noise}

\begin{figure*}[t]
    \centering
    \includegraphics[width=\textwidth]{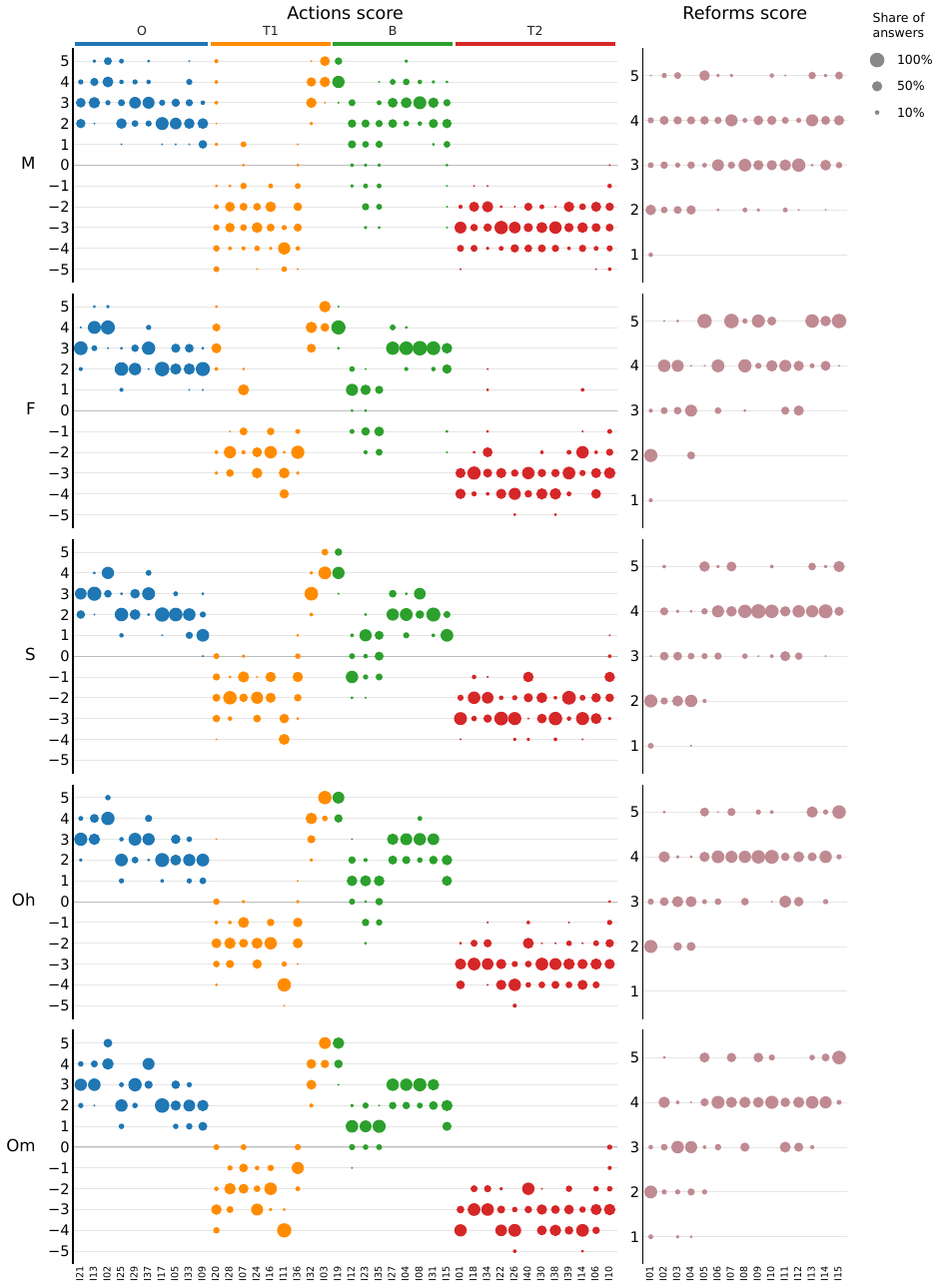}
    \caption{As Figure~\ref{fig:item_dots_all}, separately for each model (60 responses per item and model).}
    \label{fig:item_dots_models}
\end{figure*}

In Section~\ref{ssec:noise}, Figure~\ref{fig:item_dots_all}, the distribution of the scores of each item is shown for all models together. Here, in Figure~\ref{fig:item_dots_models}, it is shown for each model separately. As before, the area of a dot is proportional to the share of the responses, and the administrations are shown in the same colors as throughout the paper. For each model, the scores of each item cluster at least as closely as in Figure~\ref{fig:item_dots_all}: within one point of the model's median for the item lie 92\% (Mistral) to 99\% of its responses on the actions, and 93\% to 100\% of its responses on the reforms.

\section{Difference between Models, by Administration}\label{app:model_model_by_admins}

\begin{figure*}[t!]
    \centering
    \includegraphics[width=\textwidth]{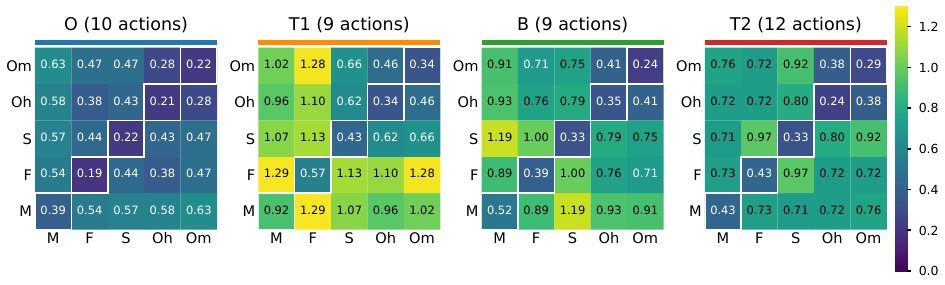}
    \caption{The mean absolute difference between the scores of two models, as Figure~\ref{fig:MM_diff_scores}, but for the scores of the actions of each administration separately: Equations~\ref{eq:diff_MM_nondiag} and~\ref{eq:diff_MM_diag} with $X$ restricted to the actions of one administration. O: Obama; T1 and T2: the first and second Trump administrations; B: Biden.}
    \label{fig:MM_diff_scores_by_admin}
\end{figure*}

In Section~\ref{ssec:model_model}, Figure~\ref{fig:MM_diff_scores}, the differences between the models' scores of the presidential actions were averaged over all administrations. Here, in Figure~\ref{fig:MM_diff_scores_by_admin}, they are shown for each administration separately. Averaged over the pairs of different models, the difference is 0.48 for the Obama actions, 0.84 for Biden, 0.96 for the first Trump administration and 0.74 for the second; the two runs of the same model differ by 0.25, 0.36, 0.52 and 0.35, respectively.

\begin{figure*}[t]
    \centering
    \begin{minipage}[t]{0.48\textwidth}
        \centering
        \includegraphics[width=\linewidth]{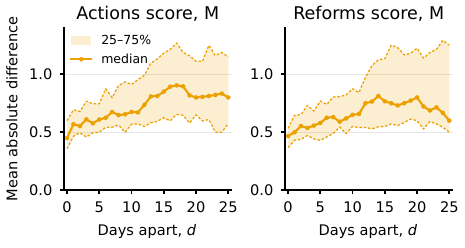}
        \caption{Drift of the scores of M (Mistral, Think mode): the median over the items of Equations~\ref{eq:drift_scores} and~\ref{eq:drift_scores_0}, and the range between their first and third quartiles.}
        \label{fig:drift_model_M}
    \end{minipage}\hfill
    \begin{minipage}[t]{0.48\textwidth}
        \centering
        \includegraphics[width=\linewidth]{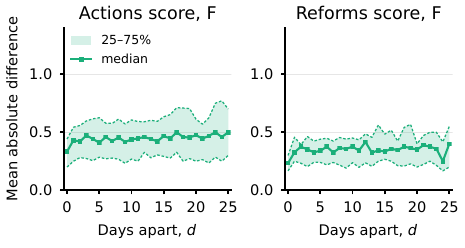}
        \caption{As Figure~\ref{fig:drift_model_M}, for F (Gemini Flash 3.6).}
        \label{fig:drift_model_F}
    \end{minipage}

    \medskip
    \begin{minipage}[t]{0.48\textwidth}
        \centering
        \includegraphics[width=\linewidth]{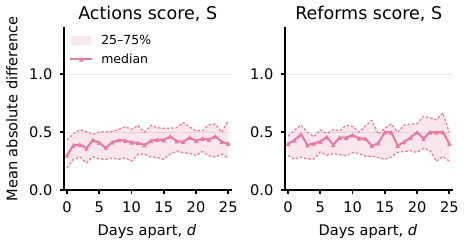}
        \caption{As Figure~\ref{fig:drift_model_M}, for S (Claude Sonnet 5, effort High).}
        \label{fig:drift_model_S}
    \end{minipage}\hfill
    \begin{minipage}[t]{0.48\textwidth}
        \centering
        \includegraphics[width=\linewidth]{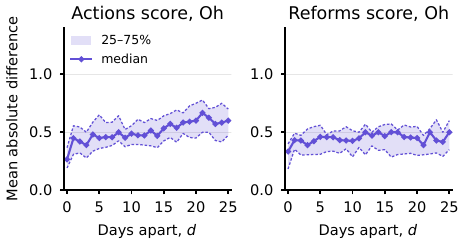}
        \caption{As Figure~\ref{fig:drift_model_M}, for Oh (Claude Opus 5, effort High).}
        \label{fig:drift_model_Oh}
    \end{minipage}

    \medskip
    \begin{minipage}[t]{0.48\textwidth}
        \centering
        \includegraphics[width=\linewidth]{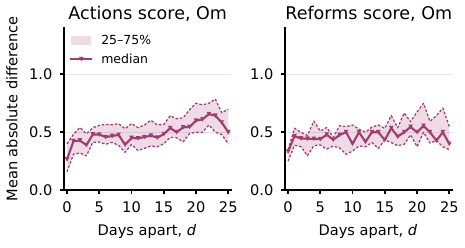}
        \caption{As Figure~\ref{fig:drift_model_M}, for Om (Claude Opus 5, effort Max).}
        \label{fig:drift_model_Om}
    \end{minipage}
\end{figure*}

\section{Drift by Model}\label{app:drift_by_model}
In Section~\ref{ssec:drift_scores}, Figure~\ref{fig:drift_models}, the drift of the scores was averaged over the items. Here, in Figures~\ref{fig:drift_model_M}--\ref{fig:drift_model_Om}, it is shown for each model separately, as the median over the items and the range between their first and third quartiles.

\section{Daily Means}\label{app:model_changes}
\begin{figure*}[t]
    \centering
    \includegraphics[width=\textwidth]{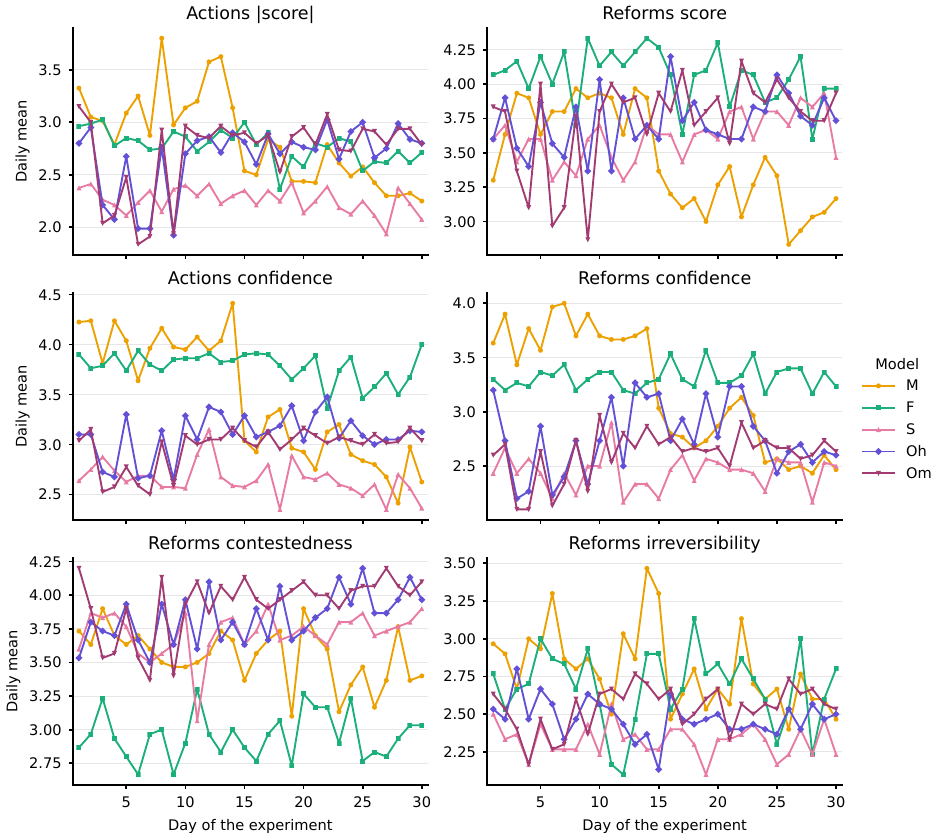}
    \caption{Daily means of each model's responses, averaged over the items and the two runs of each day: the absolute score of the actions, the score of the reforms, the confidence of both, and the contestedness and irreversibility of the reforms.}
    \label{fig:daily_means}
\end{figure*}

Besides the gradual drift, which may come from changes in public knowledge and sentiment, a chatbot can change abruptly: its routing, the model served, or any other part of the system may be updated without notice. Such changes can be seen in the daily means of the responses, averaged over the items and the two runs of each day, shown in Figure~\ref{fig:daily_means}. For the scores of the actions, the mean of the absolute value is shown, because the signed mean stays near zero for every model: positive and negative scores cancel. Three of the changes are confirmed statistically in Section~\ref{ssec:model_changes}.

\begin{figure*}[t]
  \centering
  \includegraphics[width=0.9\textwidth]{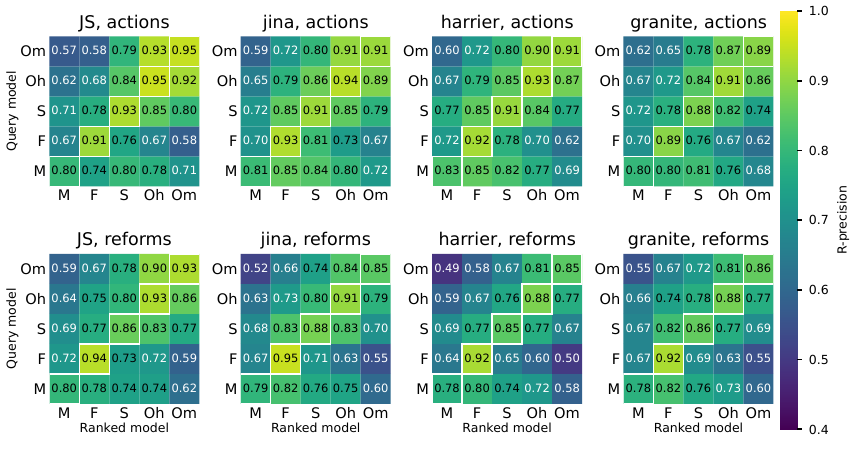}
  \caption{As Figure~\ref{fig:ret-map}, with R-precision: the share of the
  30 relevant comments among the first 30 in the ranking.}
  \label{fig:ret-rprec}
\end{figure*}

\section{Consistency of Comments: R-precision}\label{app:comments_consistency}

In Section~\ref{sssec:comments_consistency}, the consistency of the comments was measured as the quality of a retrieval task, so that any measure of ranking quality can be used. Figure~\ref{fig:ret-map} there shows the mean average precision (MAP). Here, Figure~\ref{fig:ret-rprec} shows the R-precision, which gives the same picture.

\end{document}